\documentclass[letterpaper, 10 pt, conference]{ieeeconf}  
\IEEEoverridecommandlockouts                              

\usepackage[english]{babel}

\usepackage{amsthm}
\usepackage{amsmath}
\usepackage{amssymb}
\usepackage{booktabs}
\usepackage[dvips]{graphicx}
\usepackage{multirow}
\usepackage{amsfonts}
\usepackage{enumerate}
\let\labelindent\relax
\usepackage{enumitem}
\usepackage{tabularx}
\usepackage{algorithm,algorithmic}
\usepackage{bm}
\usepackage{adjustbox}
\usepackage{xcolor}
\usepackage{siunitx}
\usepackage{mdframed}
\usepackage{adjustbox}
\usepackage{authblk}
\usepackage{color}
\usepackage{xurl}
\usepackage{subcaption}
\usepackage{hhline} 
\usepackage{cite}
\makeatletter
\let\NAT@parse\undefined
\makeatother
\usepackage{relsize}
\usepackage{float}
\usepackage{pifont}
\usepackage{hyperref}

\DeclareMathOperator*{\argmin}{arg\,min}

\title{\LARGE \bf
Graph Neural Planning and Predictive Control for Multi-Robot Communication-Constrained Unlabeled Motion Planning

}
\author{Manohari Goarin$^1$, Yang Zhou$^1$, and Giuseppe Loianno$^2$
\thanks{$^1$The authors are with New York University, NY 10012, USA. {\tt\footnotesize email: \{mg7363, yz5794\}@nyu.edu}.}
\thanks{$^2$The author is with the University of California Berkeley,
Department of Electrical Engineering and Computer Sciences,
Berkeley, CA 94720, USA. {\tt\footnotesize email: loiannog@eecs.berkeley.edu}.}
\thanks{This work was supported by the DARPA YFA Grant D22AP00156-00, the DEVCOM ARL grant SARA W911NF-24-2-0057, and the NSF CPS Grant CNS-2603416.}
}

\bstctlcite{IEEEexample:BSTcontrol}

\begin{document}
\bstctlcite{IEEEexample:BSTcontrol}

\maketitle
\thispagestyle{empty}
\pagestyle{empty}

\begin{abstract}
The multi-robot unlabeled motion planning problem of concurrently assigning robots to goals and generating safe trajectories is central in many collaborative tasks. Recent Graph Neural Network methods offer scalable decentralized solutions but rely on simplified dynamics and simulation environments, overlooking key challenges of real-world deployment such as dynamic feasibility and communication constraints. To address these gaps, we propose a hierarchical framework that combines a Graph ATtention Planner (GATP) with a decentralized Nonlinear Model Predictive Controller (NMPC). GATP provides intermediate subgoals through multi-robot cooperation, and the NMPC enforces safety under nonlinear dynamics and actuation constraints. We evaluate our framework in both simulation and real-world quadrotor experiments. Thanks to attention mechanisms and minimal communication requirements, we demonstrate improved generalization to larger teams, robustness to communication delays up to 200 ms and practical feasibility with decentralized on-board inference.
\end{abstract}


\noindent \textbf{Video}: \href{https://youtu.be/OocpY1PnUfs}{https://youtu.be/OocpY1PnUfs}

\noindent \textbf{Code}: \href{https://github.com/mgoarin7363/GATP-Graph-Attention-Planner}{https://github.com/mgoarin7363/GATP}
\section{Introduction}

In recent years, multi-robot systems have attracted significant attention due to their ability to speed up task execution compared to single robot solutions, while concurrently offering additional resilience to robot failures. Cooperative multi-robot planning has been extensively studied for exploration, surveillance, search and rescue, or warehouse automation \cite{verma2021multi, wang2025breaking}. By coordinating their actions, robots can solve these tasks with increased energy and time efficiency. The unlabeled motion planning problem offers a unifying formulation for many of these applications when the robots are homogeneous and interchangeable. It is a joint assignment and trajectory planning problem, with the objective of cooperatively reaching a set of goals while minimizing the total distance and time to travel and avoiding collisions.

While centralized methods can find optimal solutions, the computational burden of large robot teams make them impractical, motivating the development of decentralized methods.
Among recent works, learning-based methods and especially Graph Neural Networks (GNNs) have shown strong potential to solve multi-robot collaborative tasks \cite{li2020graph, ji2021decentralized,hu2023graph, gosrich2022coverage, tolstaya2020learning,zhou2022multi,zhou2022graph,tolstayamulti, zhang2022h2gnn} and notable works have applied them to the decentralized unlabeled motion planning problem \cite{khan2021large, muthusamy2024generalizability, khan2019graph, wang2023hierarchical}. GNNs are inherently decentralized through their message-passing architecture and can exploit the structural information of the team topology to learn solutions that are close to optimal while scaling to large robot teams. However, these GNN-based methods rely on simplified dynamics in simulation environments and overlook key challenges for real-world deployment such as communication constraints, trajectory smoothness, and guaranteed safety under actuation limits.
\begin{figure}[t]
    \centering
    \includegraphics[width=\columnwidth]{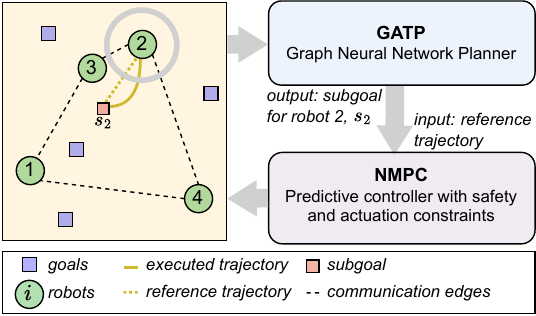}
    \caption{
    \textbf{Hierarchical Architecture for cooperative and safe unlabeled motion planning:} a Graph ATtention Planner (GATP) that exchanges information over the robot communication graph and provides subgoals; a Nonlinear Model Predictive Control (NMPC) that tracks these subgoals with safety and actuation constraints. 
    }
    \label{fig:overview}
    \vspace{-16pt}
\end{figure}

To address these challenges in unlabeled motion planning, we propose a hierarchical architecture that integrates a GNN for high-level prediction of subgoals with a decentralized Nonlinear Model Predictive Control (NMPC) for trajectory execution toward these subgoals (see Figure \ref{fig:overview}).
This hierarchy leverages the complementary strengths of both components: the GNN captures collaborative behavior through inter-robot communication to guide the robots toward their goals, and the NMPC guarantees safety and smooth motion under nonlinear dynamics and actuation constraints. The planner provides intermediate goals to the robots over a future horizon, and continuously updates them according to the robots’ position changes during NMPC execution. This dynamic-agnostic high-level planning formulation allows low frequency replanning and direct deployment in real-world scenarios. Moreover, our GNN architecture relies on minimal communication, improving robustness to communication delays.
We summarize our contributions as follows
\begin{itemize}[left=0pt]
    \item We propose a hierarchical framework combining a Graph ATtention Planner (GATP) and an NMPC for multi-robot collaborative and safe unlabeled motion planning with minimal communication.
    \item We perform an ablation study on our GNN architecture and benchmark it against Graph Convolution Networks (GCN) commonly used in prior works. We compare their coverage performance and generalization across various team sizes.
    \item We test our GATP-NMPC in simulation with $10$ quadrotors in two application scenarios, circle formation and zone coverage. We evaluate its coverage time performance under increasing communication delays and show its robustness for delays under $200$ \si{ms}.
    \item We implement a ROS~2 decentralized inference algorithm and deploy our framework in real-world experiments with $4$ quadrotors. We quantify the inference times and communication delays to demonstrate its practical feasibility.
\end{itemize}

\section{Related Works}
\vspace{-1pt}
Multi-robot motion planning is particularly challenging in the unlabeled setting, where robots are interchangeable and the joint assignment–planning problem is PSPACE-hard \cite{solovey2016hardness}. Centralized solutions exist, such as C–CAPT \cite{turpin2014capt} that employs the Hungarian algorithm for assignment \cite{kuhn1955hungarian} and constant-velocity trajectory generation with collision avoidance guarantees under certain conditions. However, they are often impractical in real-world scenarios since they rely on global knowledge of the team’s state, which is typically unavailable due to communication constraints. Moreover, they scale poorly as the number of robots increases. These limitations have motivated the development of decentralized approaches, presented in this section.

\textbf{Model-based unlabeled motion planning.}
These methods often solve the assignment problem explicitly and combine it with robot and task-dependent trajectory optimization techniques. Switching-based strategies \cite{bang2021energy, dergachev2024decentralized, turpin2014capt, hu2020convergent, panagou2019decentralized} allow robots to iteratively refine their goal selection through local interactions with their neighbors, requiring little communication, but often suffering from slow or suboptimal convergence. Other methods rely on decentralized optimization-based and auction-based algorithms \cite{lusk2020distributed, sung2020distributed, xu2024multi, morgan2016swarm}. They can achieve better assignments, but typically require numerous communication rounds before converging to an optimal solution, and degrade in restricted communication settings \cite{goarin2024graph}. In addition, explicit assignment can lead to abrupt changes in direction resulting in inefficient trajectories.

\textbf{Learning-based unlabeled motion planning.}
Learning-based methods, including reinforcement learning \cite{khan2019learning, setyawan2022cooperative, qie2019joint, wang2021multirobot, elfakharany2021end} and unsupervised learning \cite{sellers2023autonomous}, have been explored to enhance the scalability and efficiency of unlabeled motion planning. Notably, GNNs gained significant attention due to their
intrinsic ability to model and process graph-structured data,
composed of nodes and edges, using graph filters \cite{wu2020comprehensive}. Naturally decentralized, they offer an effective representation
for multi-robot systems with robot nodes and communication
edges. Through a message-passing framework, robots share information with their neighbors and make predictions locally.
They have been applied successfully to various multi-robot applications like path planning \cite{li2020graph, ji2021decentralized}, coverage \cite{hu2023graph, gosrich2022coverage, tolstayamulti}, flocking \cite{hu2023graph, tolstaya2020learning}, collaborative perception \cite{zhou2022multi}, target tracking \cite{zhou2022graph}, or exploration \cite{tolstayamulti, zhang2022h2gnn}.
By optimizing information sharing between robots, GNNs can learn near-optimal solutions with limited communication \cite{goarin2024graph}.

In the context of unlabeled motion planning, GNN-based methods have been proposed to learn end-to-end the concurrent goal assignment and trajectory planning problem \cite{khan2021large, muthusamy2024generalizability, khan2019graph, wang2023hierarchical}. Collision avoidance is typically handled either as a soft constraint within a reinforcement learning framework \cite{muthusamy2024generalizability, khan2019graph, wang2023hierarchical}, or by applying a safety filter to the GNN outputs \cite{khan2021large}. 
However, these approaches rely on simplified dynamics in simulation and directly predict velocity commands, which may cause abrupt accelerations and assume dynamic feasibility on real robots. Moreover, incorporating collision avoidance as a penalty term does not guarantee safety, while applying a post hoc safety filter can alter the network outputs and degrade performance. Finally, these GNN-based methods require multi-hop communication and $4$ to $5$ layers, which impose a heavy communication burden and make the system more vulnerable to delays in real-world settings.

Building on these works, this paper proposes a hierarchical GATP–NMPC framework that maintains trajectory optimality and safety guarantees under nonlinear dynamics and actuation limits. In addition, our GNN relies on only $2$ layers of communication which is more robust to practical delays, and is directly deployable in real-world. To the best of our knowledge, we are the first to validate our GNN-based unlabeled motion planner in real-world settings with fully decentralized, on-board inference on the robots.


\textbf{Real-world Communication challenges for GNNs.}
A major challenge when deploying GNNs on multi-robot systems is communication. While most learning-based works assume reliable message exchanges, real-world deployments are subject to communication delays, drops, and asynchronous updates \cite{gielis2022critical}. A handful of studies evaluate their GNNs in physical multi-robot experiments, and typically rely on off-board centralized inference \cite{ji2021decentralized, wang2024multi}. In contrast, the authors in \cite{blumenkamp2022framework} introduce a decentralized ROS~2 framework to execute inference directly on-board.
By testing standard velocity-planning GNN architectures, they notice that communication constraints, typically communication delays, lead to non-negligible performance degradation.
In this paper, we analyze the impact of increasing communication delays on our task performance and show that our GNN is robust to bounded delays under $\sim200$ \si{ms}.


\section{Problem Formulation}
In the following, all variables are time-dependent, but we
drop the $t$ for better clarity. We denote column vectors with bold notation like $\mathbf{x}$, matrices with capital notation like $\mathbf{A}$, and scalars with unbold notation like $d$. We consider a team of $N$ identical robots $i \in \{1, ..., N\}$ and a set of $N$ goals $ j \in \{1, ..., N\}$.
Each robot follows nonlinear dynamics:
\vspace{-1pt}
\begin{equation*}
    \forall i, \quad \mathbf{\dot{x}}_i = f(\mathbf{x}_i,\mathbf{u}_i),
\end{equation*}
with $\mathbf{x}$ the state of robot $i$, including its position $\mathbf{p}_i$  and velocity $\mathbf{v}_i$, and $\mathbf{u}_i$ its control input. 
The unlabeled motion planning objective is to compute minimum-time and collision-free trajectories from some initial positions to the goal positions as follows:
\begin{enumerate}[left=0pt]
    \item \textbf{Coverage objective.} Robots are interchangeable, not pre‑assigned to goals and need to cover all goals as fast as possible. The task is complete when each goal is within a coverage threshold $c$ of some robot:
    \vspace{-2pt}
    \[
        \forall j,\quad \min_i \left\| \mathbf{p}_j - \mathbf{p}_i \right\|_2 < c.
    \]
    \item \textbf{Pairwise safety.} For all times $t$ and robot pairs $(i,i')$,
    \vspace{-3pt}
    \[
        \left\| \mathbf{p}_i - \mathbf{p}_{i'} \right\|_2 > d_{\mathrm{safe}},
    \]
    with $d_{\mathrm{safe}}$ the safety distance to maintain.
\end{enumerate}

In a decentralized setting, robots have limited sensing and communication capabilities. We consider that each robot $i$ can sense its $P_g$ closest goals and $P_r$ closest robots and estimate their relative positions. We define its observation as
\vspace{-5pt}
\begin{equation*}
    \mathbf{o}_i = \begin{bmatrix}
        \mathbf{p}_i^\top & \mathbf{g}_{i,1}^\top \cdots \mathbf{g}_{i,P_g}^\top & \mathbf{r}_{i,1}^\top \cdots \mathbf{r}_{i,P_r}^\top \\
    \end{bmatrix}^\top,
\end{equation*} with $\mathbf{g}_{i,j}$ the relative position of goal $j$ with respect to $i$, and $\mathbf{r}_{i,i'}$ the relative position of robot $i'$ with respect to $i$. $\mathcal{O} = \{\mathbf{o}_i \mid i \in \{1,...,N\}\}$ is the set of all robots' observations.  $\mathcal{R}_i$ is the set of all $P_r$ robots sensed by robot $i$.

Additionally, a robot $i$ can communicate with its $M$ closest neighbors, and we note this neighborhood $\mathcal{N}_i$.

\section{Methodology}

\subsection{Hierarchical Planning and Control Overview} \label{sec:gnn_method}
Our hierarchical framework is depicted in Fig.~\ref{fig:overview}. From an initial set of robot and goal positions, we can model a graph connecting robot nodes with communication edges. Each robot is connected to its $M$ closest neighbors ($M=2$ on the figure). We define the multi-robot graph as $\mathcal{G} = \{\mathcal{O}, \mathbf{A}\}$. The adjacency matrix $\mathbf{A} \in \mathbb{R}^{N \times N}$ describes the topology of the graph with its coefficients being
\vspace{-1pt}
\begin{equation*}
    A_{ii'} = \begin{cases}
        1 & \text{if } i' \in \mathcal{N}_i \\
        0 & \text{otherwise}
    \end{cases}
\end{equation*}

The framework is decomposed in two main parts:
\begin{enumerate}[leftmargin=*]
    \item Our GNN-based planner GATP, applied on each robot $i$. It takes as input the local graph $\mathcal{G}_i = \{\mathbf{o}_i, \mathbf{A}_{:i}\}$ with $\mathbf{A}_{:i}$ the $i$'s column of $\mathbf{A}$, and outputs a subgoal position command $s_i$. $s_i$ is an intermediate goal point for the robot to reach within the prediction horizon $T_p$ of the planner.
    \item A Nonlinear Model Predictive Controller with a prediction horizon $T_c$. Each robot $i$'s NMPC tracks a straight linear minimum jerk reference trajectory \cite{mellinger2011minimum} to the subgoal $s_i$ while respecting hard safety and dynamic constraints. This module allows smooth and safe motion of the robots toward their subgoals.
\end{enumerate}
GATP continuously updates the NMPC reference depending on the evolution of the robots' positions during execution.

\subsection{Graph Neural Network for high-level planning}
We design our GNN architecture as a Graph Attention Network (GAT) \cite{velickovic2017graph} which improves expressiveness by learning adaptive importance weights for each node’s neighbors.
We add Multi Layer Perceptrons (MLPs) update functions between layers, allowing each node to fuse its initial embedding with aggregated neighbor information. This mechanism preserves the permutation invariance property of the GNN and enhances node differentiation.

At a robot node $i$, the GNN architecture can be decomposed into three steps: \textbf{the Encoder}, \textbf{the Attention-based Message-passing module}, and \textbf{the Decoder}. We note $\mathbf{h}_i^l$ the embedding of node $i$ at layer $l$, and $\phi_*$ MLPs with two linear layers and a $\operatorname{LeakyReLu}$ activation function in between.
\begin{enumerate}[left=0pt]
    \item \textbf{The Encoder} transforms the robots' observations into embedding vectors of size $F$
    \begin{equation}
        \mathbf{h}_i^0 \leftarrow \phi_{enc}(\Tilde{\mathbf{o}}_i),
    \end{equation}
    where $\Tilde{\mathbf{o}}_i$ is the normalized observation $\mathbf{o}_i$ with respect to the environment spatial dimension $D_w$, obtained by dividing $\mathbf{p}_i$ by $D_w$ and $\mathbf{g}_{i,j}$, $\mathbf{r}_{i,i'}$ by $2D_w$.
    \vspace{3pt}
    \item \textbf{The Attention-based Message-passing module} applies $L$ layers (or communication rounds) of 1-hop neighbor information aggregation and embedding update. At each layer $l$, first, attention coefficients $e^l_{ii'}$ are computed between node $i$ and each of its neighbors $i'$
    \begin{equation}
        \forall i' \in \mathcal{N}_i, \quad e^l_{ii'} = \mathbf{a}^l (\mathbf{W}^l \mathbf{h}^l_i, \mathbf{W}^l \mathbf{h}^l_{i'}).
    \end{equation} 
    The attention function $\mathbf{a}^l$ is a weight vector of size $2F$ followed by a $\operatorname{LeakyReLu}$. $\mathbf{W}^l$ is a weight matrix of size $F \times F$. These attention coefficients are normalized throughout the neighborhood using a softmax function
    \begin{equation}
        \forall i' \in \mathcal{N}_i, \quad \Tilde{e}^l_{ii'} = \operatorname{softmax}_{i'}(e^l_{ii'}).
    \end{equation}
    Finally, a weighted sum using the coefficients $\Tilde{e}^l_{ii'}$ followed by a nonlinear activation function combines the neighbors' embeddings
    \begin{equation}
        \Bar{\mathbf{h}}^l_i = \operatorname{tanh} \bigg(\sum_{i' \in \mathcal{N}_i} \Tilde{e}^l_{ii'} \mathbf{W}^l \mathbf{h}^l_{i'} \bigg).
    \end{equation}
    We apply multi-head attention by performing $K$ attention mechanisms in parallel. These heads $k$ are combined using the $\operatorname{max}$ function
    \begin{equation}
        \hat{\mathbf{h}}^l_i = \max_{k \in \{1,...,K\}}  \Bar{\mathbf{h}}^{l,k}_i .
    \end{equation}
    Once this message-passing is performed, we update node $i$'s embedding by combining its initial encoded embedding $\mathbf{h}^0_i$ with the aggregated information $\hat{\mathbf{h}}^l_i$
    \begin{equation}
        \mathbf{h}^{l+1}_i \leftarrow \phi_{update}(\mathbf{h}^0_i \mid\mid \hat{\mathbf{h}}^l_i).
    \label{eq:update} \end{equation}
    This fusion strategy improves expressiveness and node differentiation.
     \vspace{3pt}
    \item \textbf{The Decoder} predicts an output $\Tilde{s}_i$ for $i$ from its final embedding $\mathbf{h}^L_i$
    \begin{equation}
        \Tilde{s}_i = \operatorname{tanh} ( \phi_{dec} (\mathbf{h}^L_i) ), \quad \in [-1,1]^n.
    \end{equation}
    The Cartesian dimension $n$ is $2$ or $3$ depending on whether the environment is 2D or 3D.
    Then this output is converted into a relative subgoal command by scaling and clamping it to its maximum magnitude, i.e, its maximum desired distance $S_p$ from $i$
    \begin{equation}
        s_i = \min \bigg( 1, \frac{S_p}{\mid\mid \Tilde{s}_i \mid\mid_2} \bigg) \Tilde{s}_i,
    \end{equation}
    where $S_p = v_{max}T_p$ represents the spatial horizon of the planner, given a desired time horizon $T_p$ and a desired maximum speed $v_{max}$ of the robots.
\end{enumerate}
\subsection{Safe Predictive Control for Trajectory Execution} 
We formulate a nonlinear optimization problem over the prediction horizon $T_c$ for a robot $i$ in continuous time
\begin{subequations}
\begin{align} 
    &\argmin_{\mathbf{u}_i}&& \int_{t_0}^{t_0 + T_c} \bigg( \|\mathbf{x}_i - \mathbf{x}_{i}^*(s_i)\|_Q^2 + \|\mathbf{u}_i - \mathbf{u}_{i}^*\|_R^2 \bigg) \, dt \label{quad_cost}\\
   &\text{s.t.} \quad \forall t,  &&\dot{\mathbf{x}}_i = f(\mathbf{x}_i,\mathbf{u}_i), \label{dyn_cons}\\
   &&&\mathbf{x}_i \in \mathcal{X},~\mathbf{u}_i \in \mathcal{U}, \label{xu_cons}\\
   &&&G_{ii'}(\mathbf{x}_i,\mathbf{x}_{i'},\mathbf{u}_i) \geq 0,~ \forall i' \in \mathcal{R}_i \label{h_cons},
\end{align}
\end{subequations}
where Eq.~\eqref{quad_cost} represents the quadratic objective function minimizing the distance to the min-jerk reference trajectory $\mathbf{x}_i^*(s_i)$ guiding the robot toward its subgoal $s_i$ with a maximum desired speed $v_{max}$, and the reference control inputs $\mathbf{u}_i^*$. Eq.~\eqref{dyn_cons} is the nonlinear dynamics of the robot, Eq.~\eqref{xu_cons} the state and input constraints, and Eq.~\eqref{h_cons} safety constraints maintaining the safety distance $d_{safe}$ between $i$ and the closest robots it can sense $\{i' \in \mathcal{R}_i\}$. In this paper, we choose to formulate $G_{ii'}$ as Control Barrier Function (CBF) constraints which can provide strong safety guarantees under actuation constraints \cite{goarin2024decentralized}. The NMPC prediction horizon $T_c$ is lower or equal than GATP prediction horizon $T_p$.
\vspace{5pt}
\section{Experimental Setup}
\subsection{GATP Training Setup}
GATP is trained with imitation learning. The centralized expert employs the Hungarian Algorithm \cite{kuhn1955hungarian} to assign the robots to the goals minimizing the total distance traveled. The optimal subgoals $s^*_i$ are calculated by discretizing straight trajectories toward the assigned goals with a step size $S_p$. During training, the robots follow simple simulation dynamics under a mixed GATP/expert policy
\begin{equation*}
    \forall i, \forall t, \quad \mathbf{p}_i(t+1) = \begin{cases}
        s_i(t) & \text{with a probability $\beta$} \\
        s^*_i(t) & \text{with a probability 1-$\beta$}
    \end{cases}
\end{equation*}
We adopt a scheduled sampling scheme, gradually increasing $\beta$ to replace expert subgoals with the GATP predictions. The planning spatial horizon is set at $S_p = 4$ m, which defines the maximum magnitude of the predicted subgoals. Since the GNN outputs are normalized, different horizons can still be applied at test time. The choice of $S_p$ was empirically tuned to balance convergence speed and stability near the goals: larger horizons lead to oscillations around the targets, while smaller horizons result in slower learning and convergence.

We build a dataset of $10000$ graphs of $N=10$ agents with random initial and goal positions in an environment of $20 \times 20$ meters, and rollout trajectories of $N_t = 70$ timesteps. We train our GNN with centralized inference using the Deep Graph Library \cite{wang2019deep} to minimize the mean squared error between the normalized predicted subgoals $\Tilde{s}_i$ and the normalized expert subgoals $\Tilde{s}^*_i$, summed over trajectories and averaged across robots per batch. The loss is formulated as
\begin{equation*}
    \mathcal{L} = \frac{1}{NB}\sum_{i} \sum_t  w_i(t)\mid\mid  \Tilde{s}_i(t) - \Tilde{s}^*_i(t) \mid\mid^2_2,
\end{equation*}
with $B$ the batch size. $w_i(t)$ is an adaptive weight that increases when the error is small, to refine the predictions' accuracy and stabilize the planning behavior around the goal locations. It is obtained as
\begin{equation*}
    w_i(t) = 1 + \alpha_1 \exp \bigg(-\alpha_2 \mid\mid  \Tilde{s}_i(t) - \Tilde{s}^*_i(t) \mid\mid_2 \bigg),
\end{equation*}
where the parameters $\alpha_1$ and $\alpha_2$ are hand-tuned.

In the context of restricted communication scenarios in the real world, and to reduce the GNN sensitivity to possible delays during deployment, we limit the message-passing module to 1-hop aggregations, $L=2$ layers, and $M=2$ maximum neighbors each robot can communicate with. We restrict the observation of the robots to $P_g = 5$ closest goals and $P_r = 3$ closest robots, use $K=3$ heads of attention, and a feature dimension $F = 64$. 
In the following subsections, the training runs and comparisons were conducted using a batch size of $B=200$ graphs and 80 epochs on a 12th-generation Intel
CPU I9-12900H. We increase the probability $\beta$ incrementally from $0.5$ to $1.0$ every $20$ epochs, and fix the loss parameters $\alpha_1 = 5.0$ and $\alpha_2 = 4.0$. The learning rate is set at $6\mathrm{e}{-4}$ and $1\mathrm{e}{-4}$ on the last $20$ epochs.

\subsection{Quadrotor dynamics}
In our experiments, we demonstrate and deploy our framework with quadrotors. The state and control inputs of a quadrotor $i$ can be described as
\begin{equation*}
    \mathbf{x}_i = \begin{bmatrix}
    \mathbf{p}_i^\top & \mathbf{v}_i^\top & \mathbf{q}_i^\top & \boldsymbol{\omega}_i^\top
\end{bmatrix}^\top, \mathbf{u}_i=\begin{bmatrix}
    u_{i0} & u_{i1} & u_{i2} & u_{i3}
\end{bmatrix}^\top,
\end{equation*}
where $\mathbf{p}_i\in \mathbb{R}^3$ and $\mathbf{v}_i\in \mathbb{R}^3$ are respectively the position and linear velocity of the quadrotor in the inertial frame, $\mathbf{q}_i\in \mathbb{R}^4$ the rotation in quaternions from the quadrotor's body frame to the inertial frame, $\boldsymbol{\omega}_i\in \mathbb{R}^3$ the angular velocity in the body frame, and $\{u_{ik}\in \mathbb{R}, k\in[0,...,3]\}$ the motor thrusts of the quadrotor. The dynamic equations are as presented in \cite{saviolo2023learning} and can be written in a control-affine form as follows
\begin{equation*}
    \dot{\mathbf{x}}_i = f'(\mathbf{x}_i) + g(\mathbf{x}_i) \mathbf{u}_i.
\end{equation*}
The safety constraints in Eq.~\eqref{h_cons} from the NMPC formulation are based on Exponential Control Barrier Functions (ECBFs) which can provide forward invariance guarantees of the safe set for the quadrotors' higher-order dynamics. We refer the readers to \cite{goarin2024decentralized}
for further details.

\section{Results and Analysis in Simulation and Real-World}
\begin{figure*}[th] 
    \centering
    \begin{minipage}{0.48\textwidth}
        \centering
        \includegraphics[width=\linewidth]{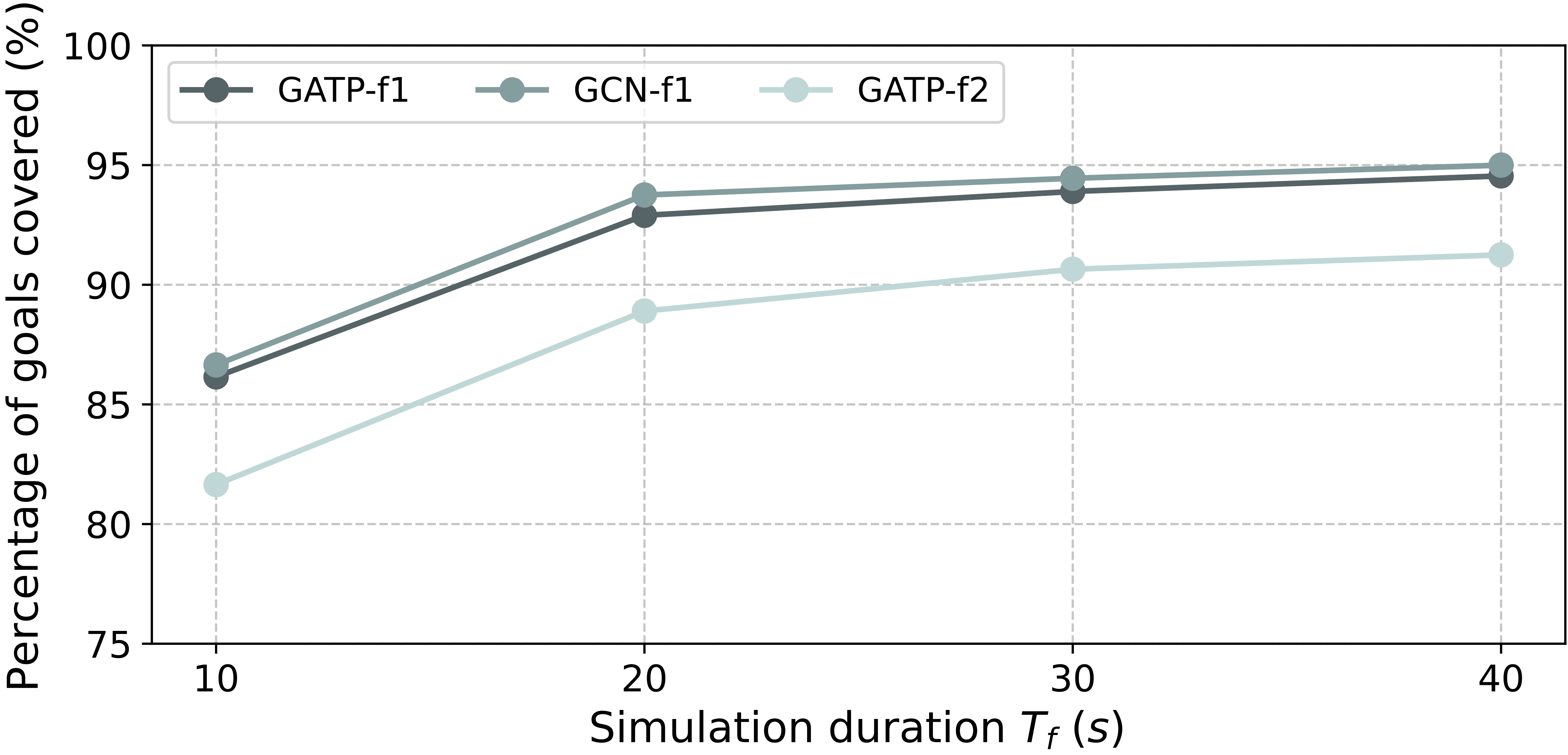}
        \caption{GATP coverage performance analysis with $10$ robots.}
        \label{fig:gatp_perf1}
    \end{minipage}
    \hfill
    \begin{minipage}{0.49\textwidth}
        \centering
        \vspace{10pt}
        \includegraphics[width=\linewidth]{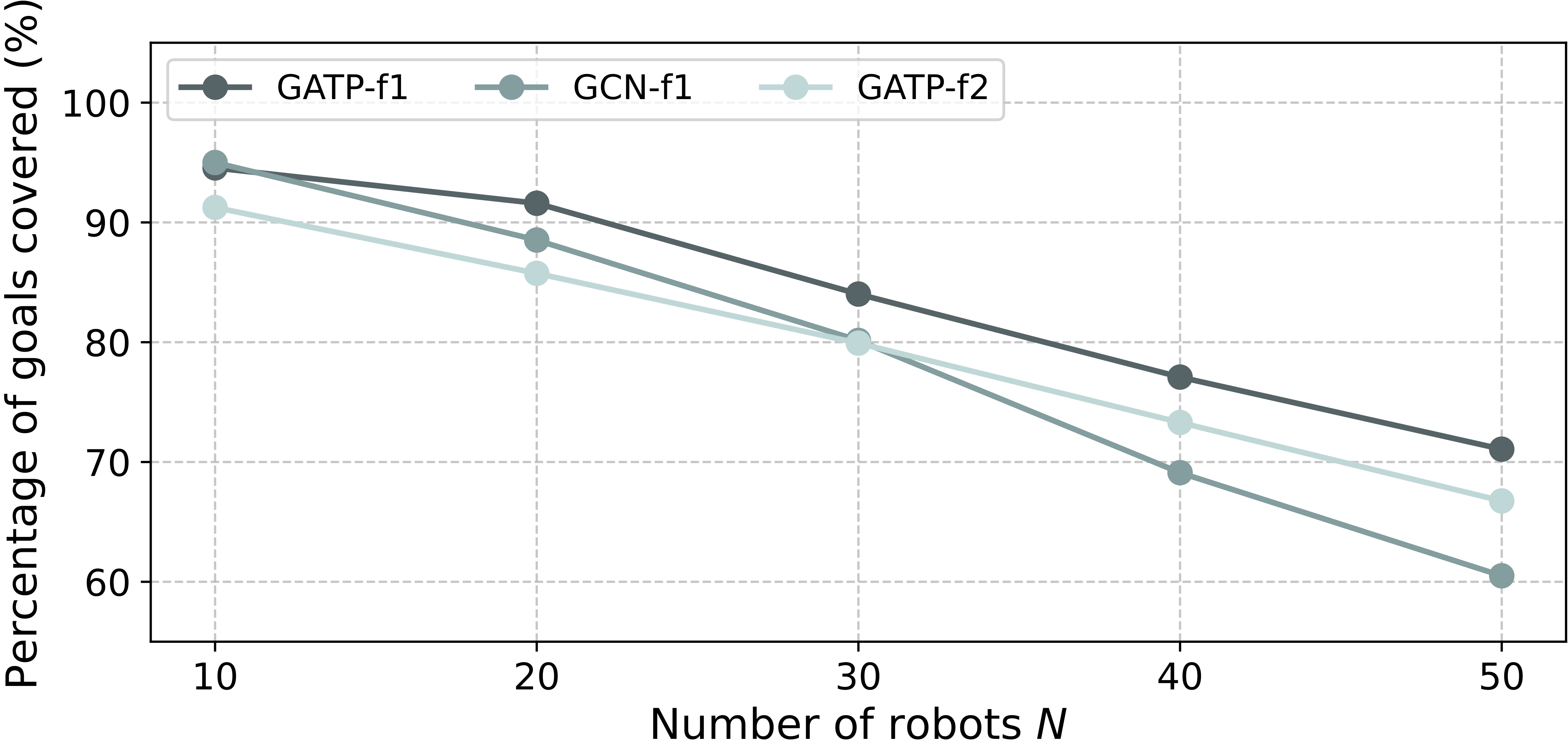}
        \caption{GATP generalization analysis to larger teams with a simulation duration of $T_f = 40$s.}
        \label{fig:gatp_perf2}
    \end{minipage}
    \vspace{-5pt}
\end{figure*}
\subsection{GATP coverage performance analysis}

First, we evaluate the planning performance of GATP using point-like robots and show the benefits of attention and our MLP fusion strategy (Eq.~\eqref{eq:update}) on coverage performance and generalization to larger teams. We define coverage performance as the percentage of goals reached in a given time $T_f$. The spatial horizon is still fixed at $S_p = 4$ m. We report in Figs.~\ref{fig:gatp_perf1} and \ref{fig:gatp_perf2} the total coverage percentage over $200$ scenarios, for different simulation durations $T_f$ from $10$ to $40$ s and different team sizes from $N=10$ to $50$ robots. We test 3 different GNN architectures for comparison:
\begin{itemize} 
    \item \textbf{GATP-f1}, our architecture as presented in section \ref{sec:gnn_method}, with $f_1$ the update function from Eq.~\eqref{eq:update}: $f_1 (i,l) = \phi_{update}(\mathbf{h}^0_i \mid\mid \hat{\mathbf{h}}^l_i)$.
    \item \textbf{GCN-f1}, a Graph Convolutional Network as employed in \cite{khan2021large,muthusamy2024generalizability,khan2019graph}, similar to GATP-f1 but with $2$ layers of 1-hop convolutional filters instead of attention layers.
    \item \textbf{GATP-f2}, a variation of GATP-f1 that uses a different update function $f_2(i,l) = \mathbf{h}^l_i + \phi_{update}(\hat{\mathbf{h}}^l_i)$ that combines the aggregated information with the previous node embedding instead of the initial encoded embedding. This function was used in \cite{muthusamy2024generalizability}.
\end{itemize}

First, we analyze the coverage performance on $10$ robots with increasing simulation durations (Figure~\ref{fig:gatp_perf1}). GCN-f1 and GATP-f1 achieve comparable results. With a limited time of $T_f = 10$ s, GATP-f1 reaches $86.15 \%$ of the goals compared to $86.65\%$ for GCN-f1, despite relying on only 1-hop neighbor aggregation and 2 communication rounds. In $T_f = 40$ s, GATP-f1 reaches $94.55 \%$ of coverage against $95.00\%$ for GCN-f1. These results exhibit the ability of GNNs to learn efficient heuristics through optimized information exchange. Both architectures outperform GATP-f2 which completes at most $81.45\%$ in $10$ s and $91.25 \%$ in $40$ s. The greater performance of GCN-f1 and GATP-f1 can be attributed to the update function $f_1$ that re-injects the initial node embedding after each layer via a learnable MLP. This strategy enhances node differentiation and reduces assignment conflicts.

Second, we analyze the generalization of the GNN architectures to larger teams of up to $50$ robots. For lower computation cost and faster training, the GNNs are trained with $10$ robots but are directly transferrable to larger team sizes thanks to the decentralized nature of GNNs. On Figure~\ref{fig:gatp_perf2}, we analyze the evolution of the coverage percentage as we increase the number of robots for a fixed simulation duration of $40$ s. Overall, the more robots we add, the more the GNNs degrade in performance because of the small number of robots used for training. However, GATP-f1 outperforms both GCN-f1 and GATP-f2. Although GCN-f1 demonstrated similar performance to GATP-f1 with $10$ robots, GATP-f1 and GATP-f2 are more generalizable to larger teams with a smaller decrease in performance. From $10$ to $50$ robots, GATP-f1 experiences a performance drop of $23.5\%$, compared to $24.5\%$ for GATP-f2 and $34.5\%$ for GCN-f1. We conclude that attention mechanisms improve generalization. By dynamically weighting neighbors’ information, the network can capture more complex interactions and account for the varying importance of different neighbors.


\subsection{GATP-NMPC performance analysis under increasing communication delays}
\begin{figure*}[th] 
    \centering
    \begin{minipage}{0.48\textwidth}
        \centering
    \includegraphics[width=\linewidth]{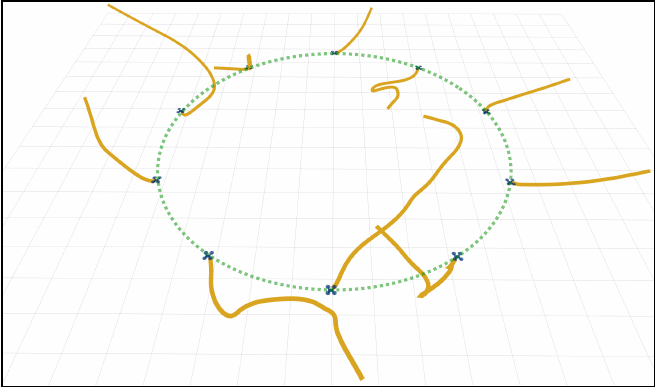}
        \subcaption{Circle Formation}
    \end{minipage}
    \hfill
    \begin{minipage}{0.48\textwidth}
        \centering
        \includegraphics[width=\linewidth]{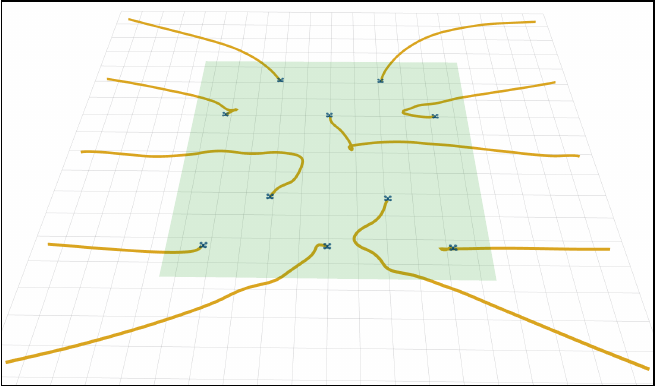}
        \subcaption{Zone Coverage}
    \end{minipage}
    \caption{Simulation with $10$ quadrotors for a circle formation task and a coverage task. The green circle is the desired circle to form, and the green area is the environment zone to cover. The quadrotors are in blue, and their trajectories are yellow.}
    \label{fig:sim_illustration}
\end{figure*}
\begin{figure}[t!] 
    \centering
    \begin{minipage}{0.51\columnwidth}
        \centering
        \includegraphics[width=\linewidth]{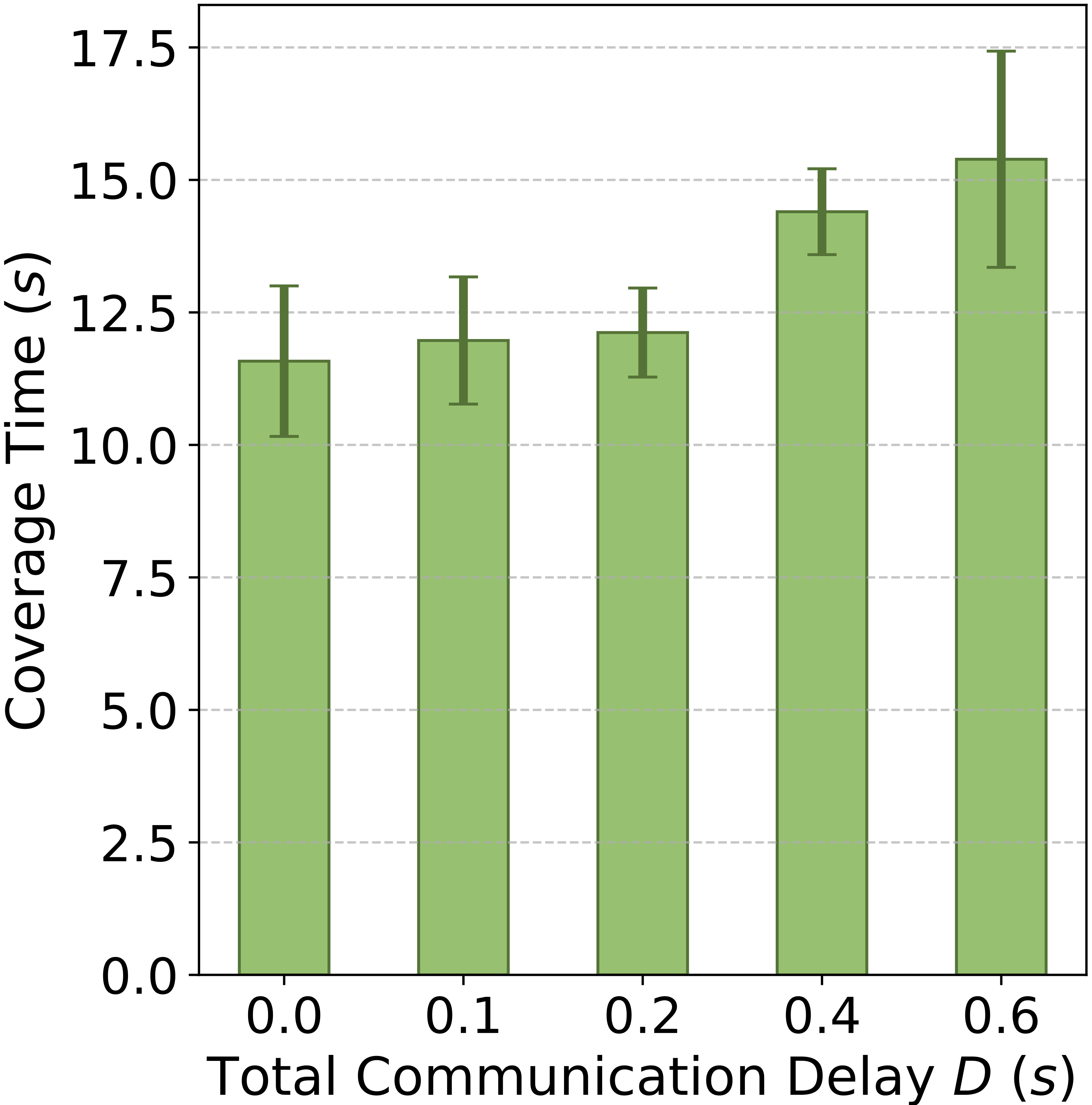}
        \subcaption{Circle Formation}
    \end{minipage}
    \hfill
    \begin{minipage}{0.46\columnwidth}
        \centering
        \includegraphics[width=\linewidth]{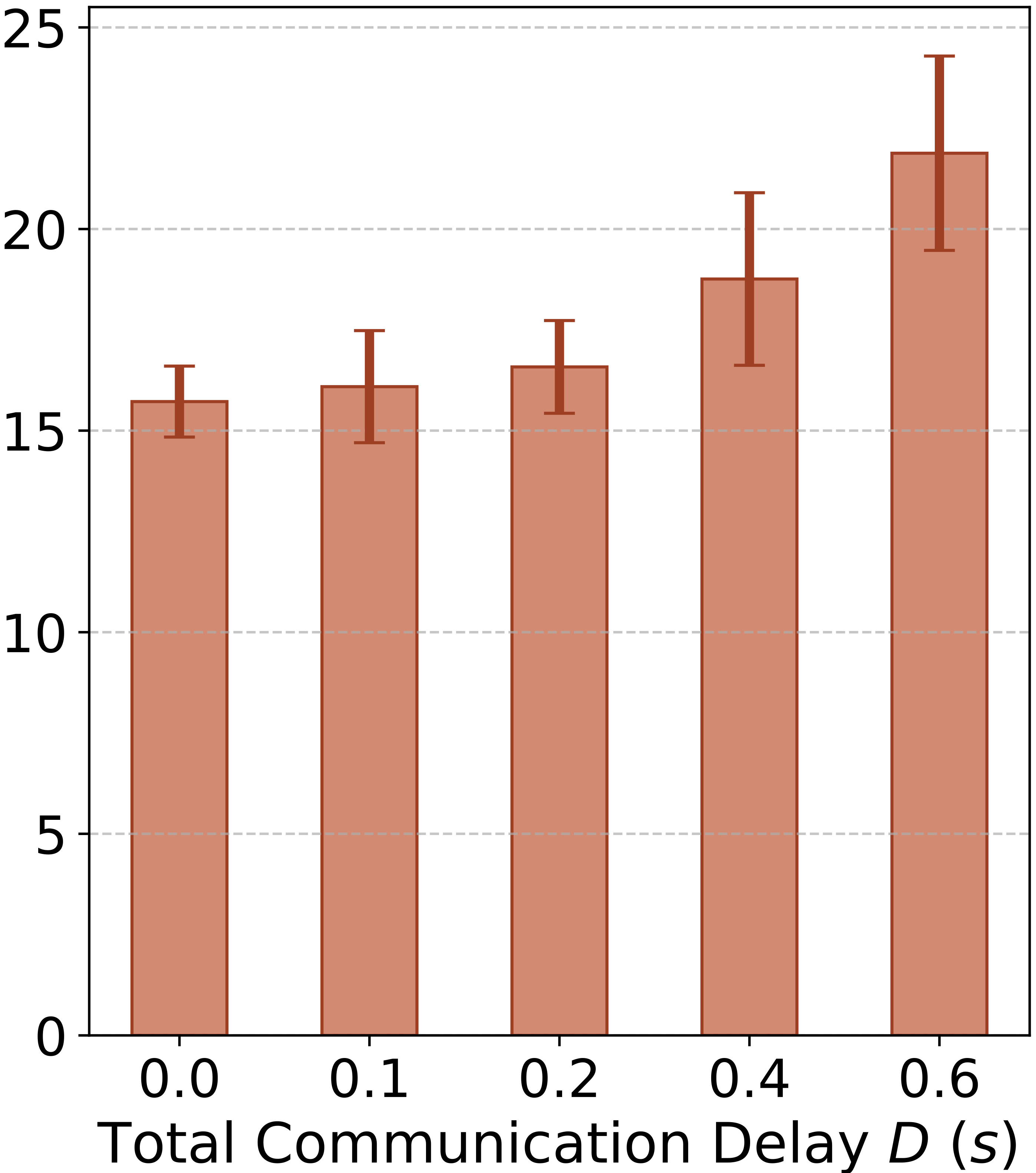}
        \subcaption{Zone Coverage}
    \end{minipage}
    \caption{Performance analysis under increasing communication delays on both tasks.}
    \label{fig:com}
    \vspace{-10pt}
\end{figure}
In this section, we test our hierarchical GATP-NMPC  in simulation with quadrotors. We run $10$ robot nodes in a ROS~2 environment, each running its own NMPC but with a centralized GATP node, only for this simulation experiment, to avoid overloading our computer with parallel inferences. The planner runs at $2$ \si{Hz} and the NMPC at $100$ \si{Hz}. We set the GATP time horizon at $T_p= 2$ s and the desired maximum speed of the robots at $v_{max} = 2$ \si{m/s}, which is equivalent to a planning spatial horizon of $S_p = 4$ \si{m}. The desired safety distance between robots is $d_{safe}= 0.5$ \si{m}. We use acados \cite{Verschueren2021} to solve the NMPC with SQP-RTI and a prediction horizon of $T_c = 1.5$ \si{s}.

We demonstrate the efficacy of our framework in two example tasks:
(a) a \textbf{circle formation task}, where robots start from a set of random positions and must form a circle, and (b) a \textbf{ zone coverage task}, where robots must uniformly cover a given zone of the environment. Two simulation runs illustrate these tasks in Figure \ref{fig:sim_illustration}. Additional examples for different tasks can be found in the video.
The performance metric for these experiments is the \textbf{coverage time}, defined as the total time required for the team to reach all goals, with a coverage threshold of $c = 0.2$ m per goal.

First, the GATP–NMPC framework successfully guides the quadrotors to their goal locations in both tasks, producing smooth and safe trajectories through the seamless integration of the graph-based planner with the NMPC. The high-level planner leverages local information to minimize the distance traveled by the robots, helping to deconflict the multi-robot system and facilitate NMPC feasibility.

In the real world, significant delays may arise from inter-robot communication, which the message-passing module depends on, and can lead to planning updates based on outdated observations. To model this effect, we assume that all robots experience the same communication delay $d$ per communication round (i.e., GNN layer). The total communication delay is then $D = Ld$. We test different values of  $D$ and analyze its impact on coverage time performance for both tasks. The inference time also contributes to the total delay but is negligible compared to the communication effect in practice.
We run each task $10$ times with identical initial and goal positions, and report the mean and standard deviation of the coverage time in Fig.~\ref{fig:com}. Variations across runs are due to additional uncertainties related to ROS~2 processes and numerical approximations in the NMPC.
In both scenarios, we notice some performance degradation for large communication delays $D=0.4$ \si{s} and $D=0.6$ \si{s}. However, the framework is robust to delays under $0.2$ s, with an average increase of coverage time of $0.44$ \si{s} ($3.8$\%) on the circle formation task, and $0.86$ \si{s} ($5.5$\%) on the zone coverage task. The performance degradation becomes impactful when the delay is comparable to the planning update interval of $0.5$ \si{s}. At such delays, the outputs are too outdated to remain valid for effective planning. With $D = 0.6$ \si{s}, the coverage time increases by $31.8\%$ in the circle formation task and by $39.2\%$ in the zone coverage task. The effect is more pronounced in the last scenario, where robots must travel longer distances and thus accumulate larger errors over time.

In conclusion, our GATP architecture with $L=2$ layers for high-level, low-frequency planning is robust enough to bounded communication delays of up to $0.2$ s (i.e., $d = 0.1$ \si{s} per layer). Using more layers (e.g., $4$–$5$ as in previous works \cite{khan2021large,muthusamy2024generalizability, khan2019graph}) could potentially improve planning performance, but would also make the system significantly more sensitive to delays, since with $0.1$ s per layer the total communication latency would already reach $D = 0.4$ to $0.5$ \si{s}.

\subsection{System Implementation and Deployment in Real-World}
\begin{figure*}[t!] 
    \begin{minipage}{0.6\textwidth}
        \includegraphics[width=\linewidth]{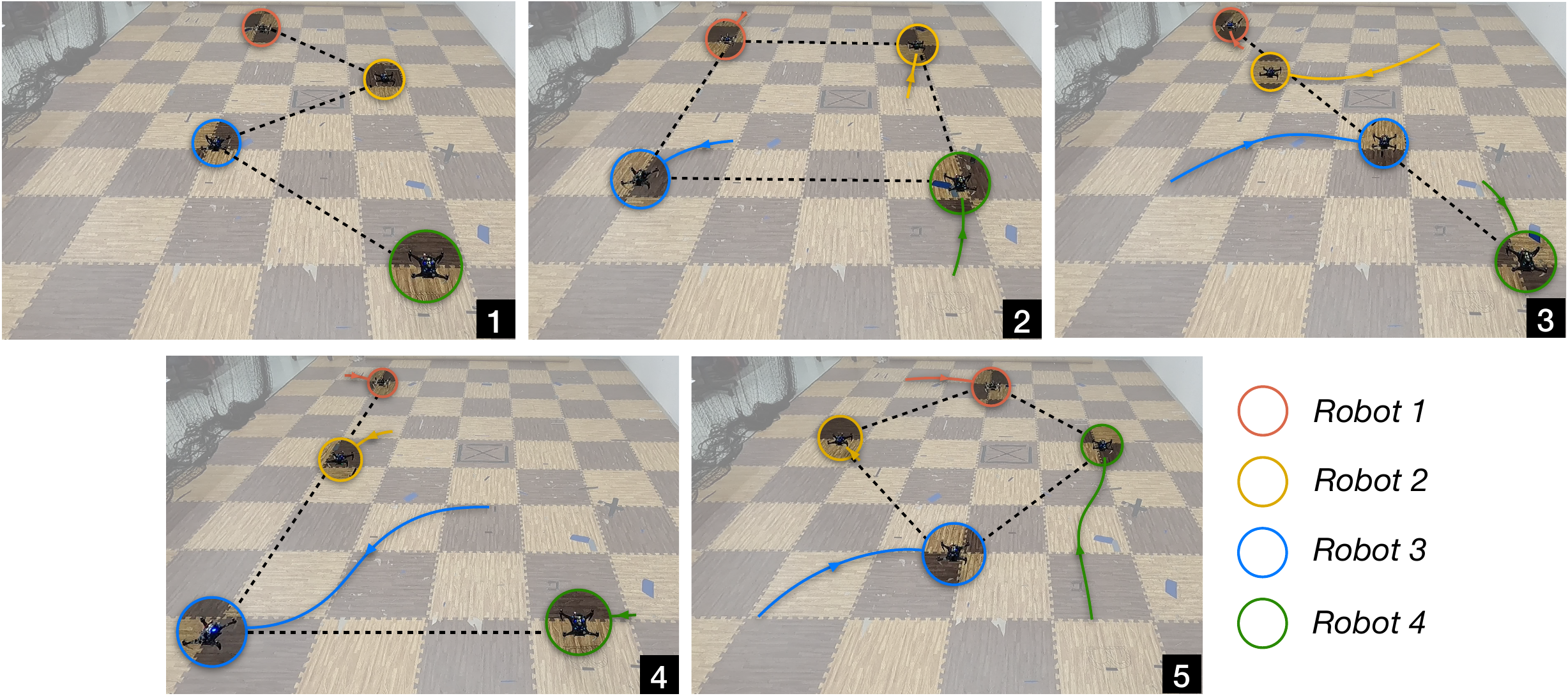}
        \subcaption{Without obstacles}
        \label{fig:rw_exp1}
    \end{minipage}
    \begin{minipage}{0.38\textwidth}
        \includegraphics[width=\linewidth]{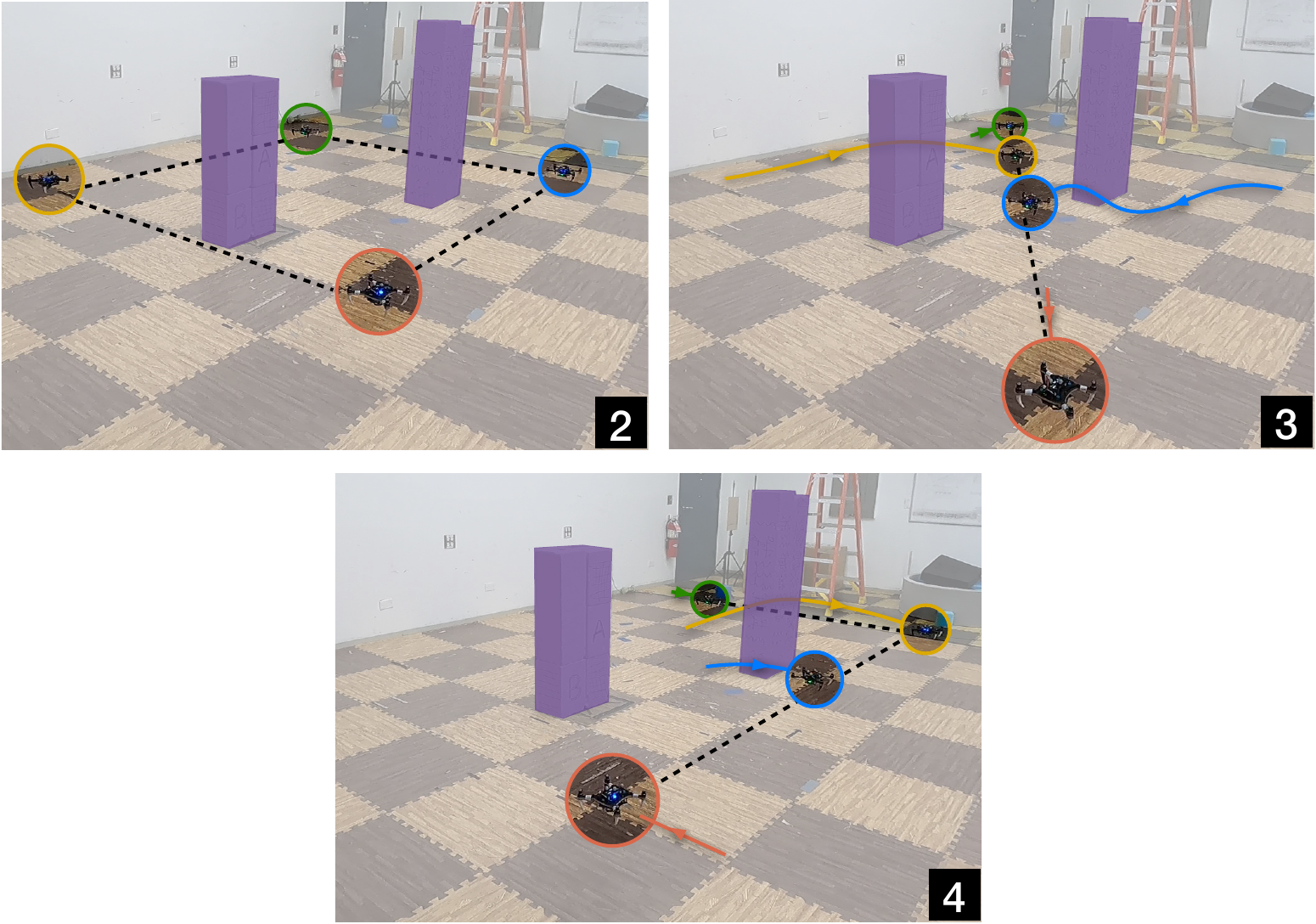}
        \subcaption{With obstacles}
        \label{fig:rw_exp2}
    \end{minipage}
    \caption{Real-world Experiments. Each circle represents a robot, the arrow line its trajectory, and the dashed line a desired shape. Color red is used for Robot $1$, yellow for Robot $2$, blue for Robot $3$, green for Robot $4$, and purple for obstacles.}
    \label{fig:rw_exp}
\end{figure*}

We deploy our solution in an indoor \(\SI{10}{m}\times\SI{6}{m}\times\SI{4}{m}\) testbed with Vicon\footnote{\url{https://www.vicon.com/}} localization and $4$ custom quadrotors based on~\cite{LoiannoRAL2017} and equipped with Qualcomm\textsuperscript{\textregistered} Snapdragon\textsuperscript{TM} VOXL\textsuperscript{\textregistered}~2\footnote{\url{https://www.modalai.com/products/voxl-2?variant=39914779836467}}. We demonstrate the applicability of our GATP–NMPC framework in a formation task, where robots sequentially form different shapes (see Figure \ref{fig:rw_exp} and the attached multimedia material). Taking into account our environmental dimensions, we set the desired maximum speed of the robots at $v_{max} = 0.5 $ \si{m/s}, the planner time horizon at $T_p = 1.5$ \si{s}, and the NMPC horizon $T_c = 1$ \si{s}. Each robot senses its $P_g = 4$ closest goals and $P_r = 2$ closest neighbors.
Both the planner and the controller run onboard each quadrotor at $1$ \si{Hz} and $160$ \si{Hz}, respectively. $1$ \si{Hz} was sufficient for this experimental setup given the small flying area and the robots' velocity.

{
\setlength{\textfloatsep}{-5pt}
\begin{algorithm}[h]
\caption{GATP inference ROS~2 node on quadrotor $i$}
\label{algo:inference}
\begin{algorithmic}[1] 
\WHILE{$t< T_f$}
    \STATE Get observation $\mathbf{o}_i$
    \STATE Run encoder
    \FOR{$l=1$ to $L$}
        \STATE Wait for all neighbors' messages
        \IF{all messages received}
            \STATE Aggregate messages
            \STATE Update embedding
        \ENDIF
    \ENDFOR
    \STATE Run decoder
    \STATE Send subgoal command to NMPC 
\ENDWHILE
\end{algorithmic}
\end{algorithm}
}

We implement a decentralized inference as in Algorithm~\ref{algo:inference}. For each layer, robots wait for all neighbors' messages before proceeding. GATP replans concurrently with the NMPC until the experiment ends.
\setlength{\skip\footins}{7pt}

Our framework transfers directly to the real world thanks to its hierarchical design: the NMPC accounts for nonlinear quadrotor dynamics with actuation constraints while GATP uses position‑based observations and commands that are dynamic‑agnostic, facilitating sim-to-real transfer and generalization across platforms.
For all tested formation shapes, quadrotors successfully reach the goals safely as shown on Figure \ref{fig:rw_exp1} and in the video. In this experiment, the coverage time performance is similar in simulation and the real world.

We further validate our framework in the presence of obstacles (Figure \ref{fig:rw_exp2}). 
Two obstacles were placed along the transitions between shapes 2→3 and 3→4. During 2→3, the robots have to deviate significantly from their reference trajectories to avoid the obstacles. GATP re-planned accordingly, providing a new global solution in which the yellow and blue quadrotors swapped positions compared to the obstacle-free case (Figure \ref{fig:rw_exp1}). In contrast, during 3→4, obstacle avoidance did not alter the optimal solution, and GATP maintained the same plan.

Finally, we measure the average inference times and communication delays during the experiment (Table \ref{table}). The total inference time of GATP is only about $1$ \si{ms}, thanks to its compact architecture and the use of ONNX Runtime\footnote{\url{https://onnxruntime.ai/}}. In contrast, the average communication delay per layer is around $26$ \si{ms}, confirming that delays are mainly due to communication rather than computation. We also observe some delay variability between robots, with a total GATP time per robot ranging from $51$ \si{ms} to $134$ \si{ms}. A $134$ \si{ms} of planning time implies that GATP could be run at a maximum frequency of roughly $7$ Hz, which is compatible with our low-frequency subgoal planning formulation. Moreover, this maximum total delay falls within the $0.1$–$0.2$ \si{s} range tested in simulation, where coverage performance was shown to remain mostly unaffected. This justifies the absence of noticeable performance loss between simulation and real-world.

\renewcommand{\arraystretch}{1.3}
\begin{table}[t]
\centering
\caption{Inference times and Communication delays}
\begin{tabular}{|l|c|c|} 
\hline
\hhline{|===|} 
\textbf{Metric} & \multicolumn{2}{c|}{\textbf{Time (ms)}} \\
\hline
Average GATP inference time (comm. excluded) & \multicolumn{2}{c|}{1.03} \\
\hline
Average communication delay per layer & \multicolumn{2}{c|}{26.04} \\
\hline
\multirow{4}{*}{Total GATP time per robot} 
 & Robot 1& 62.88 \\
 & Robot 2& 134.29 \\
 & Robot 3& 106.90 \\
 & Robot 4& 51.23 \\
\hhline{|===|} 
\end{tabular}
\vspace{-5pt}
\label{table}
\end{table}

\section{Conclusion}
In this work, we introduced a novel hierarchical approach that combines a Graph ATtention Planner (GATP) with a decentralized Nonlinear Model Predictive Controller (NMPC) for collaborative and safe unlabeled motion planning under nonlinear dynamics and limited communication. In simulation, we demonstrated improved generalization to larger teams thanks to attention mechanisms, and robustness of our two‑layer design against delays up to \SI{200}{ms}. We deployed decentralized on‑board inference on quadrotors, highlighting practical feasibility and effective sim-to-real transfer.

To strengthen coordination between hierarchical modules, future work will investigate coupling planning and control during training by incorporating dynamics knowledge and collision avoidance constraints into the GNN learning process.
In addition, we plan to examine more realistic communication conditions to better understand their impact, such as asynchronous updates and variable delays, and extend our real-world validation to larger multi-robot teams.



\bibliographystyle{IEEEtran}
\bibliography{IEEEabrv,references}

@article{solovey2016hardness,
  title={On the hardness of unlabeled multi-robot motion planning},
  author={Solovey, Kiril and Halperin, Dan},
  journal={The International Journal of Robotics Research},
  volume={35},
  number={14},
  pages={1750--1759},
  year={2016},
  publisher={SAGE Publications Sage UK: London, England}
}

@inproceedings{bang2021energy,
  title={Energy-optimal goal assignment of multi-agent system with goal trajectories in polynomials},
  author={Bang, Heeseung and Beaver, Logan E and Malikopoulos, Andreas A},
  booktitle={29th Mediterranean Conference on Control and Automation (MED)},
  pages={1228--1233},
  year={2021}
}

@inproceedings{dergachev2024decentralized,
  title={Decentralized Unlabeled Multi-agent Navigation in Continuous Space},
  author={Dergachev, Stepan and Yakovlev, Konstantin},
  booktitle={International Conference on Interactive Collaborative Robotics},
  pages={186--200},
  year={2024},
  organization={Springer}
}

@article{turpin2014capt,
  title={Capt: Concurrent assignment and planning of trajectories for multiple robots},
  author={Turpin, Matthew and Michael, Nathan and Kumar, Vijay},
  journal={The International Journal of Robotics Research},
  volume={33},
  number={1},
  pages={98--112},
  year={2014},
  publisher={SAGE Publications Sage UK: London, England}
}

@article{hu2020convergent,
  title={Convergent multiagent formation control with collision avoidance},
  author={Hu, Jinwen and Zhang, Houxin and Liu, Lu and Zhu, Xiaoping and Zhao, Chunhui and Pan, Quan},
  journal={IEEE Transactions on Robotics},
  volume={36},
  number={6},
  pages={1805--1818},
  year={2020},
  publisher={IEEE}
}

@article{lusk2020distributed,
  title={A distributed pipeline for scalable, deconflicted formation flying},
  author={Lusk, Parker C and Cai, Xiaoyi and Wadhwania, Samir and Paris, Aleix and Fathian, Kaveh and How, Jonathan P},
  journal={IEEE Robotics and Automation Letters},
  volume={5},
  number={4},
  pages={5213--5220},
  year={2020},
  publisher={IEEE}
}

@article{sung2020distributed,
  title={Distributed assignment with limited communication for multi-robot multi-target tracking},
  author={Sung, Yoonchang and Budhiraja, Ashish Kumar and Williams, Ryan K and Tokekar, Pratap},
  journal={Autonomous robots},
  volume={44},
  number={1},
  pages={57--73},
  year={2020},
  publisher={Springer}
}

@article{panagou2019decentralized,
  title={Decentralized goal assignment and safe trajectory generation in multirobot networks via multiple Lyapunov functions},
  author={Panagou, Dimitra and Turpin, Matthew and Kumar, Vijay},
  journal={IEEE Transactions on Automatic Control},
  volume={65},
  number={8},
  pages={3365--3380},
  year={2019},
  publisher={IEEE}
}

@article{xu2024multi,
  title={Multi-robot task allocation and path planning with maximum range constraints},
  author={Xu, Gang and Wu, Yuchen and Tao, Sheng and Yang, Yifan and Liu, Tao and Huang, Tao and Wu, Huifeng and Liu, Yong},
  journal={arXiv preprint arXiv:2409.06531},
  year={2024}
}

@article{morgan2016swarm,
  title={Swarm assignment and trajectory optimization using variable-swarm, distributed auction assignment and sequential convex programming},
  author={Morgan, Daniel and Subramanian, Giri P and Chung, Soon-Jo and Hadaegh, Fred Y},
  journal={The International Journal of Robotics Research},
  volume={35},
  number={10},
  pages={1261--1285},
  year={2016},
  publisher={SAGE Publications Sage UK: London, England}
}

@article{goarin2024graph,
  title={Graph neural network for decentralized multi-robot goal assignment},
  author={Goarin, Manohari and Loianno, Giuseppe},
  journal={IEEE Robotics and Automation Letters},
  volume={9},
  number={5},
  pages={4051--4058},
  year={2024},
  publisher={IEEE}
}

@article{kuhn1955hungarian,
  title={The Hungarian method for the assignment problem},
  author={Kuhn, Harold W},
  journal={Naval research logistics quarterly},
  volume={2},
  number={1-2},
  pages={83--97},
  year={1955},
  publisher={Wiley Online Library}
}

@inproceedings{khan2019learning,
  title={Learning safe unlabeled multi-robot planning with motion constraints},
  author={Khan, Arbaaz and Zhang, Chi and Li, Shuo and Wu, Jiayue and Schlotfeldt, Brent and Tang, Sarah Y and Ribeiro, Alejandro and Bastani, Osbert and Kumar, Vijay},
  booktitle={IEEE/RSJ International Conference on Intelligent Robots and Systems (IROS)},
  pages={7558--7565},
  year={2019}
}

@article{setyawan2022cooperative,
  title={Cooperative multi-robot hierarchical reinforcement learning},
  author={Setyawan, Gembong Edhi and Hartono, Pitoyo and Sawada, Hideyuki},
  journal={International Journal of Advanced Computer Science and Applications},
  volume={13},
  number={9},
  year={2022},
  publisher={Science and Information (SAI) Organization Limited}
}

@article{qie2019joint,
  title={Joint optimization of multi-UAV target assignment and path planning based on multi-agent reinforcement learning},
  author={Qie, Han and Shi, Dianxi and Shen, Tianlong and Xu, Xinhai and Li, Yuan and Wang, Liujing},
  journal={IEEE access},
  volume={7},
  pages={146264--146272},
  year={2019},
  publisher={IEEE}
}

@article{wang2021multirobot,
  title={Multirobot coordination with deep reinforcement learning in complex environments},
  author={Wang, Di and Deng, Hongbin},
  journal={Expert Systems with Applications},
  volume={180},
  pages={115128},
  year={2021},
  publisher={Elsevier}
}

@inproceedings{sellers2023autonomous,
  title={Autonomous multi-robot allocation and formation control for remote sensing in environmental exploration},
  author={Sellers, Timothy and Lei, Tingjun and Rogers, Huston and Carruth, Daniel W and Luo, Chaomin},
  booktitle={Autonomous Systems: Sensors, Processing, and Security for Ground, Air, Sea, and Space Vehicles and Infrastructure 2023},
  volume={12540},
  pages={250--266},
  year={2023},
  organization={SPIE}
}

@article{elfakharany2021end,
  title={End-to-end deep reinforcement learning for decentralized task allocation and navigation for a multi-robot system},
  author={Elfakharany, Ahmed and Ismail, Zool Hilmi},
  journal={Applied Sciences},
  volume={11},
  number={7},
  pages={2895},
  year={2021},
  publisher={MDPI}
}

@article{khan2021large,
  title={Large scale distributed collaborative unlabeled motion planning with graph policy gradients},
  author={Khan, Arbaaz and Kumar, Vijay and Ribeiro, Alejandro},
  journal={IEEE Robotics and Automation Letters},
  volume={6},
  number={3},
  pages={5340--5347},
  year={2021},
  publisher={IEEE}
}

@article{muthusamy2024generalizability,
  title={Generalizability of Graph Neural Networks for Decentralized Unlabeled Motion Planning},
  author={Muthusamy, Shreyas and Owerko, Damian and Kanatsoulis, Charilaos I and Agarwal, Saurav and Ribeiro, Alejandro},
  journal={arXiv preprint arXiv:2409.19829},
  year={2024}
}

@article{khan2019graph,
  title={Graph policy gradients for large scale unlabeled motion planning with constraints},
  author={Khan, Arbaaz and Kumar, Vijay and Ribeiro, Alejandro},
  journal={arXiv preprint arXiv:1909.10704},
  year={2019}
}

@article{wang2023hierarchical,
  title={Hierarchical relational graph learning for autonomous multirobot cooperative navigation in dynamic environments},
  author={Wang, Ting and Du, Xiao and Chen, Mingsong and Li, Keqin},
  journal={IEEE Transactions on Computer-Aided Design of Integrated Circuits and Systems},
  volume={42},
  number={11},
  pages={3559--3570},
  year={2023},
  publisher={IEEE}
}

@inproceedings{li2020graph,
  title={Graph neural networks for decentralized multi-robot path planning},
  author={Li, Qingbiao and Gama, Fernando and Ribeiro, Alejandro and Prorok, Amanda},
  booktitle={IEEE/RSJ international conference on intelligent robots and systems (IROS)},
  pages={11785--11792},
  year={2020}
}

@inproceedings{ji2021decentralized,
  title={Decentralized, unlabeled multi-agent navigation in obstacle-rich environments using graph neural networks},
  author={Ji, Xuebo and Li, He and Pan, Zherong and Gao, Xifeng and Tu, Changhe},
  booktitle={IEEE/RSJ International Conference on Intelligent Robots and Systems (IROS)},
  pages={8936--8943},
  year={2021}
}

@article{hu2023graph,
  title={Graph soft actor--critic reinforcement learning for large-scale distributed multirobot coordination},
  author={Hu, Yifan and Fu, Junjie and Wen, Guanghui},
  journal={IEEE transactions on neural networks and learning systems},
  year={2023},
  publisher={IEEE}
}

@inproceedings{gosrich2022coverage,
  title={Coverage control in multi-robot systems via graph neural networks},
  author={Gosrich, Walker and Mayya, Siddharth and Li, Rebecca and Paulos, James and Yim, Mark and Ribeiro, Alejandro and Kumar, Vijay},
  booktitle={IEEE International Conference on Robotics and Automation (ICRA)},
  pages={8787--8793},
  year={2022}
}

@inproceedings{tolstayamulti,
  title={Multi-robot coverage and exploration using spatial graph neural networks.},
  author={Tolstaya, Ekaterina and Paulos, James and Kumar, Vijay and Ribeiro, Alejandro},
  booktitle={IEEE/RSJ International Conference on Intelligent Robots and Systems (IROS)},
  year={2021},
  pages={8944--8950}
}

@article{zhang2022h2gnn,
  title={H2GNN: Hierarchical-hops graph neural networks for multi-robot exploration in unknown environments},
  author={Zhang, Hao and Cheng, Jiyu and Zhang, Lin and Li, Yibin and Zhang, Wei},
  journal={IEEE Robotics and Automation Letters},
  volume={7},
  number={2},
  pages={3435--3442},
  year={2022},
  publisher={IEEE}
}

@inproceedings{zhou2022graph,
  title={Graph neural networks for decentralized multi-robot target tracking},
  author={Zhou, Lifeng and Sharma, Vishnu D and Li, Qingbiao and Prorok, Amanda and Ribeiro, Alejandro and Tokekar, Pratap and Kumar, Vijay},
  booktitle={IEEE International Symposium on Safety, Security, and Rescue Robotics (SSRR)},
  pages={195--202},
  year={2022}
}

@article{zhou2022multi,
  title={Multi-robot collaborative perception with graph neural networks},
  author={Zhou, Yang and Xiao, Jiuhong and Zhou, Yue and Loianno, Giuseppe},
  journal={IEEE Robotics and Automation Letters},
  volume={7},
  number={2},
  pages={2289--2296},
  year={2022},
  publisher={IEEE}
}

@inproceedings{tolstaya2020learning,
  title={Learning decentralized controllers for robot swarms with graph neural networks},
  author={Tolstaya, Ekaterina and Gama, Fernando and Paulos, James and Pappas, George and Kumar, Vijay and Ribeiro, Alejandro},
  booktitle={Conference on robot learning},
  pages={671--682},
  year={2020},
  organization={PMLR}
}

@article{wu2020comprehensive,
  title={A comprehensive survey on graph neural networks},
  author={Wu, Zonghan and Pan, Shirui and Chen, Fengwen and Long, Guodong and Zhang, Chengqi and Yu, Philip S},
  journal={IEEE transactions on neural networks and learning systems},
  volume={32},
  number={1},
  pages={4--24},
  year={2020},
  publisher={IEEE}
}

@article{gielis2022critical,
  title={A critical review of communications in multi-robot systems},
  author={Gielis, Jennifer and Shankar, Ajay and Prorok, Amanda},
  journal={Current robotics reports},
  volume={3},
  number={4},
  pages={213--225},
  year={2022},
  publisher={Springer}
}

@inproceedings{blumenkamp2022framework,
  title={A framework for real-world multi-robot systems running decentralized GNN-based policies},
  author={Blumenkamp, Jan and Morad, Steven and Gielis, Jennifer and Li, Qingbiao and Prorok, Amanda},
  booktitle={IEEE International Conference on Robotics and Automation (ICRA)},
  pages={8772--8778},
  year={2022}
}

@inproceedings{wang2024multi,
  title={Multi-Robot Obstacle-Avoidance Formation Based on Graph Neural Networks and Imitation Learning},
  author={Wang, Yu and Zhou, Zongtan and Dai, Wei and Guo, Ce and Zhu, Pengming and Liu, Peng},
  booktitle={China Automation Congress (CAC)},
  pages={5499--5504},
  year={2024}
}

@article{verma2021multi,
  title={Multi-robot coordination analysis, taxonomy, challenges and future scope},
  author={Verma, Janardan Kumar and Ranga, Virender},
  journal={Journal of intelligent \& robotic systems},
  volume={102},
  number={1},
  pages={10},
  year={2021},
  publisher={Springer}
}

@article{wang2025breaking,
  title={Breaking the Hierarchy: Taxonomies and Survey on Multi-robot Integrated Task and Motion Planning},
  author={Wang, Hanfu and Ye, Weibin and Wang, Jingchuan and Chen, Weidong},
  journal={Authorea Preprints},
  year={2025},
  publisher={Authorea}
}

@inproceedings{
velickovic2017graph,
title={Graph Attention Networks},
author={Petar Veličković and Guillem Cucurull and Arantxa Casanova and Adriana Romero and Pietro Liò and Yoshua Bengio},
booktitle={International Conference on Learning Representations},
year={2018},
}

@INPROCEEDINGS{goarin2024decentralized,
  author={Goarin, Manohari and Li, Guanrui and Saviolo, Alessandro and Loianno, Giuseppe},
  booktitle={IEEE International Conference on Robotics and Automation (ICRA)}, 
  title={Decentralized Nonlinear Model Predictive Control for Safe Collision Avoidance in Quadrotor Teams with Limited Detection Range}, 
  year={2025},
  volume={},
  number={},
  pages={5387-5393}
}

@article{saviolo2023learning,
  title={Learning quadrotor dynamics for precise, safe, and agile flight control},
  author={Saviolo, Alessandro and Loianno, Giuseppe},
  journal={Annual Reviews in Control},
  volume={55},
  pages={45--60},
  year={2023},
  publisher={Elsevier}
}

@article{LoiannoRAL2017,
	Author = {G. Loianno and C. Brunner and G. McGrath and V. Kumar},
	Journal = {IEEE Robotics and Automation Letters},
	Month = {April},
	Number = {2},
	Pages = {404-411},
	Title = {Estimation, Control, and Planning for Aggressive Flight With a Small Quadrotor With a Single Camera and IMU},
	Volume = {2},
	Year = {2017}}

@Article{Verschueren2021,
  Title                    = {acados -- a modular open-source framework for fast embedded optimal control},
  Author                   = {Robin Verschueren and Gianluca Frison and Dimitris Kouzoupis and Jonathan Frey and Niels van Duijkeren and Andrea Zanelli and Branimir Novoselnik and Thivaharan Albin and Rien Quirynen and Moritz Diehl},
  Journal                  = {Mathematical Programming Computation},
  Year                     = {2021},
}

@inproceedings{mellinger2011minimum,
  title={Minimum snap trajectory generation and control for quadrotors},
  author={Mellinger, Daniel and Kumar, Vijay},
  booktitle={2011 IEEE international conference on robotics and automation},
  pages={2520--2525},
  year={2011},
  organization={IEEE}
}

@inproceedings{wang2019deep,
  title={Deep graph library: Towards efficient and scalable deep learning on graphs},
  author={Wang, Minjie Yu},
  booktitle={ICLR workshop on representation learning on graphs and manifolds},
  year={2019}
}

@IEEEtranBSTCTL{IEEEexample:BSTcontrol,
  CTLuse_article_number     = "yes",
  CTLuse_paper              = "yes",
  CTLuse_forced_etal        = "yes",
  CTLmax_names_forced_etal  = "1",
  CTLnames_show_etal        = "1",
  CTLuse_alt_spacing        = "yes",
  CTLalt_stretch_factor     = "4",
  CTLdash_repeated_names    = "yes",
  CTLname_format_string     = "{f.~}{vv~}{ll}{, jj}",
  CTLname_latex_cmd         = "",
  CTLuse_comma              = "yes", 
  CTLuse_numbers            = "yes",
  CTLcompress               = "yes"
}

\end{document}